%% file: Template.tex
\documentclass{article}
\usepackage{spconf,graphicx,hyperref}

\usepackage{booktabs}
\usepackage{multirow}
\usepackage{amsmath}
\usepackage{amssymb}

\usepackage{float}

\title{GDCNet: Generative Discrepancy Comparison Network for Multimodal~Sarcasm~Detection}
\name{Shuguang Zhang$^{1,2}$, Junhong Lian$^{1,2}$, Guoxin Yu$^{1,2,3}$, Baoxun Xu$^{4}$, Xiang Ao$^{1,2}$
\sthanks{
Corresponding author (aoxiang@ict.ac.cn).
The research work is supported by National Key R\&D Plan No. 2022YFC3303305, the National Natural Science Foundation of China under Grant (No. 62576333, U2436209), the Strategic Priority Research Program of the Chinese Academy of Sciences under Grant No. XDB0680201, Beijing Natural Science Foundation JQ25015, the Innovation Funding of ICT, CAS under Grant No. E461060.}}

\address{
  $^{1}$ \resizebox{\linewidth}{!}{State Key Laboratory of AI Safety, Institute of Computing Technology~(ICT), Chinese Academy of Sciences~(CAS)}\\
  $^{2}$ University of Chinese Academy of Sciences, CAS\\
  $^{3}$ Pengcheng Laboratory \qquad 
  $^{4}$ Shenzhen Stock Exchange
}
\begin{document}
\ninept
\maketitle
\begin{abstract}
Multimodal sarcasm detection (MSD) aims to identify sarcasm within image–text pairs by modeling semantic incongruities across modalities. 
Existing methods often exploit cross-modal embedding misalignment to detect inconsistency but struggle when visual and textual content are loosely related or semantically indirect. While recent approaches leverage large language models (LLMs) to generate sarcastic cues, the inherent diversity and subjectivity of these generations often introduce noise. 
To address these limitations, we propose the Generative Discrepancy Comparison Network (GDCNet). This framework captures cross-modal conflicts by utilizing descriptive, factually grounded image captions generated by Multimodal LLMs (MLLMs) as stable semantic anchors. 
Specifically, GDCNet computes semantic and sentiment discrepancies between the generated objective description and the original text, alongside measuring visual-textual fidelity. These discrepancy features are then fused with visual and textual representations via a gated module to adaptively balance modality contributions. 
Extensive experiments on MSD benchmarks demonstrate GDCNet's superior accuracy and robustness, establishing a new state-of-the-art on the MMSD2.0 benchmark.
\end{abstract}
\begin{keywords}
Multimodal sarcasm detection, discrepancy generation, multimodal representation enhancement
\end{keywords}

\input{body/introduction}
\input{body/methodology}
\input{body/experiments}

\input{body/conclusion}

\vfill\pagebreak


\bibliographystyle{IEEEbib}
\bibliography{refs}

\end{document}

%% file: body/introduction.tex
\section{Introduction}
\label{sec:intro}

Sarcasm is a linguistic phenomenon in which the surface meaning of an utterance diverges significantly from the speaker's intended communicative message. It is commonly employed for humor, critique, and subtle social commentary~\cite{gibbs2007irony}, making it a potent tool for communication. The surge of multimodal social-media content has subsequently driven the advancement of Multimodal Sarcasm Detection~(MSD) as an emerging research frontier. Transitioning from text-only to multimodal contexts considerably complicates this task, as the interplay between images and text often generates irony that transcends the meaning conveyed by either modality in isolation.

\begin{figure}[t]
    \centering
    \includegraphics[width=0.95\linewidth]{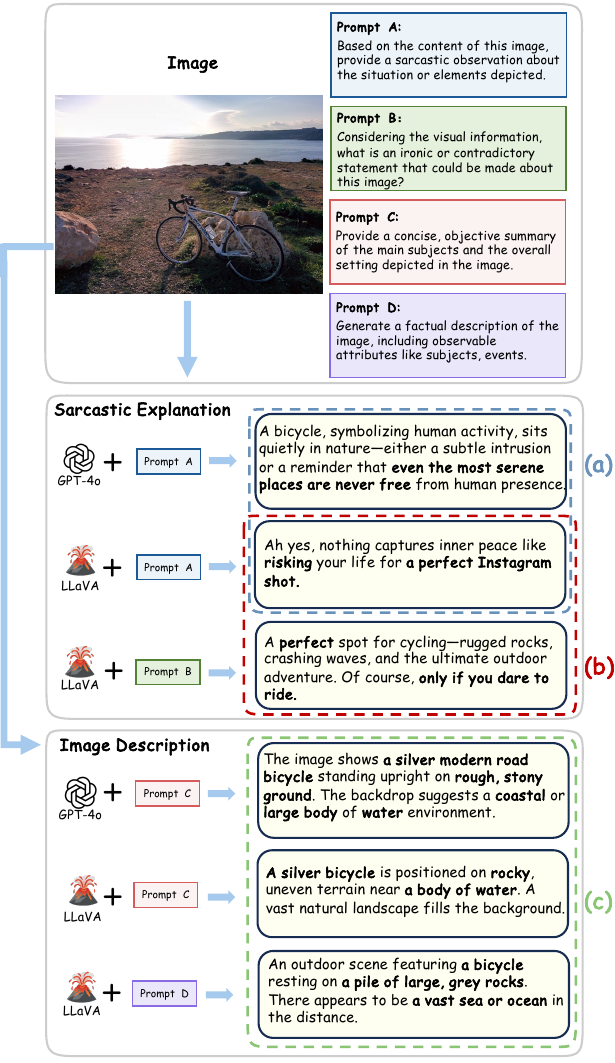}
    \caption{LLMs outputs on the same image: sarcastic explanations diverge across models and prompts, whereas factual descriptions remain stable, highlighting their potential as reliable semantic anchors.}
    \label{fig:llm_variation}
\end{figure}

MSD fundamentally relies on identifying the incongruity between images and text. Prior studies have tackled this challenge by aligning multimodal representations~\cite{cai2019multi}, leveraging techniques such as attention mechanisms~\cite{wang2020multimodal}, graph neural networks~\cite{liang2022multi}, external knowledge~\cite{liu2022sarcasm}, and dynamic routing~\cite{tian2023dynamic}. Despite these advances, existing methods~\cite{wang2020multimodal, liang2022multi, liu2022sarcasm, tian2023dynamic} still struggle with out-of-distribution generalization and often rely on superficial cues~\cite{qin2023mmsd2}. Relying solely on cross-modal embedding inconsistencies captures broad misalignments but often misses the subtle ironic cues crucial for accurate sarcasm detection, particularly when image–text alignment is weak.

Recent successes in Large Language Models~(LLMs) and their multimodal extensions~(MLLMs)~\cite{zhu2023minigpt, liu2024llavanext} have introduced extensive world knowledge and cross-modal reasoning capabilities, providing renewed impetus for the MSD task. These approaches generally exploit prompts to guide MLLMs in generating sarcastic explanations or signals for contextual data augmentation~\cite{chen-etal-2024-cofipara}. Such methods help alleviate the limitations of traditional cross-modal embedding mismatches, but they often overlook a fundamental challenge in sarcasm detection: the inherent diversity of representation perspectives. As established in cognitive psychology~\cite{gibbs2007irony}, this diversity stems from variations in human cognition and interpretation. 
The issue becomes even more pronounced in the MSD task, because distinct visual elements can contribute to generating varied ironic texts through culturally-dependent interpretations~\cite{farabi2024survey}. 
As illustrated in Fig.~\ref{fig:llm_variation}, different MLLMs (Fig.~\ref{fig:llm_variation}(a)) or even the same model with different prompts (Fig.~\ref{fig:llm_variation}(b)) generate highly divergent sarcastic expressions for the same image.
Motivated by this observation, we propose to utilize MLLMs as objective cross-modal semantic connectors rather than subjective sarcasm generators.

In this paper, we propose a framework named \textbf{G}enerative \textbf{D}iscrepancy \textbf{C}omparison \textbf{Net}work~(GDCNet). Inspired by the success of MLLMs in image captioning~\cite{liu2024improved}, we apply the MLLM to generate descriptive and factually grounded image captions, which serve as a cross-modality semantic bridge within our framework. These descriptions not only preserve visual semantics but also provide consistent and reliable semantic anchors for effective comparison with the associated textual content, as illustrated in Fig.~\ref{fig:llm_variation}(c).
Specifically, within GDCNet, we first employ an MLLM to produce objective image descriptions that serve as stable semantic references. Based on these descriptions and the corresponding original text, we introduce a \textbf{G}enerative \textbf{D}iscrepancy \textbf{R}epresentation \textbf{M}odule~(GDRM), which captures discrepancies in semantic and sentiment dimensions between texts, as well as image-text consistency at the representation level. To further enhance sarcasm detection performance, a gated fusion module is employed to integrate these multi-dimensional discrepancy representations with the original visual and textual features, thereby improving both the robustness and accuracy of the final classification.

Our main contributions are summarized as follows:
\begin{itemize}
\item We present GDCNet, a novel framework for MSD that leverages factual-grounded generated image descriptions as the semantic anchor to robustly quantify incongruity across visual and textual modalities.
\item We propose a GDRM to extract key representations for MSD by comparing semantic and sentiment differences between generated image descriptions and the original text, as well as assessing image-text fidelity.
\item Extensive experiments on widely-used benchmarks demonstrate that our GDCNet achieves significant improvements in detection accuracy and establishes a new state-of-the-art on the MMSD2.0 dataset.
\end{itemize}

%% file: body/methodology.tex
\section{Methodology}

\begin{figure*}[t]
\centering
\includegraphics[width=0.9\textwidth]{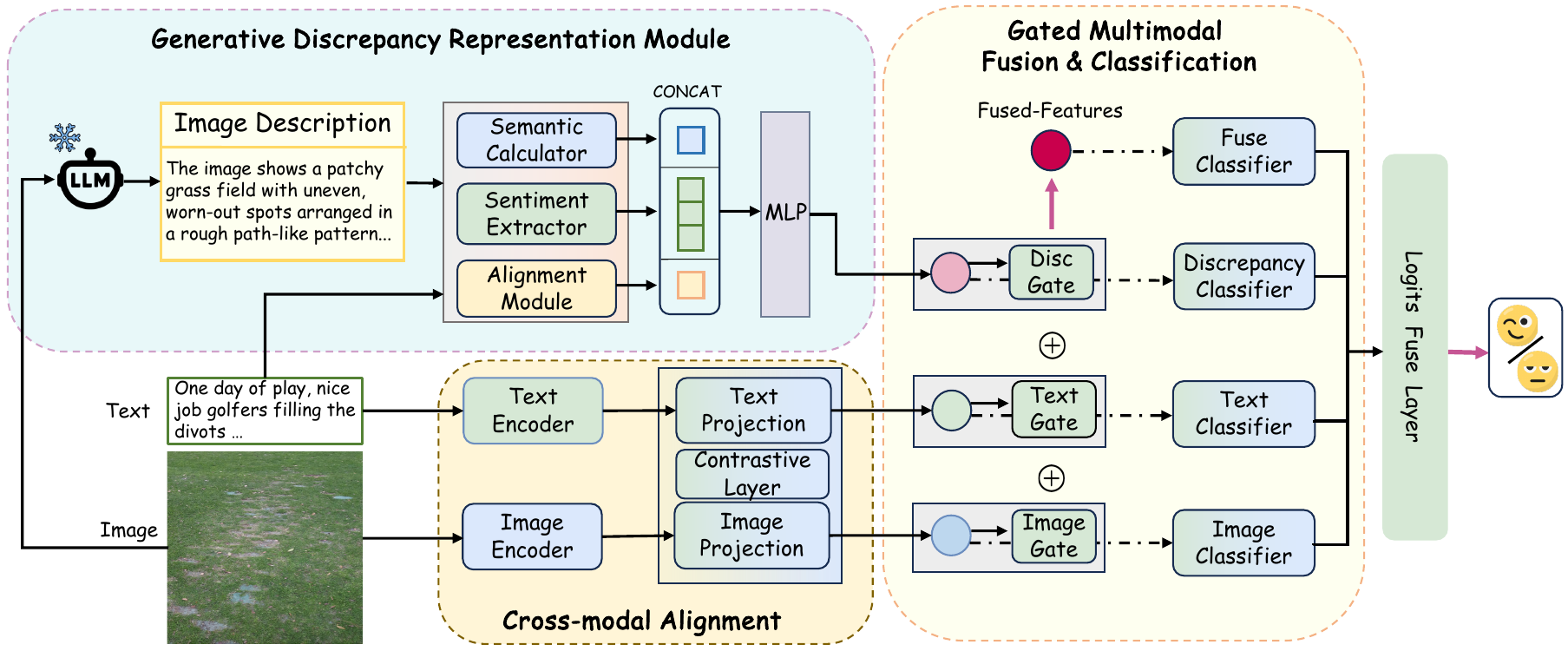}
\caption{The Architecture of GDCNet. The Gated Multimodal Fusion \& Classification module integrates discrepancy ($F_D$), text ($F_T$), and image ($F_I$) features to produce the fused representation ($F_{\text{fused}}$).}
\label{fig:framework}
\end{figure*}

\subsection{Problem Formulation}
Given an image-text pair $(I, T)$, the multimodal sarcasm detection task aims to classify it as sarcastic or non-sarcastic. 
Formally, we seek to learn a mapping $f: (I, T) \mapsto y$, with $y \in \{0, 1\}$. Here, $y=1$ indicates sarcastic and $y=0$ otherwise. 
The core challenge lies in identifying cross-modal incongruities, which often manifest as subtle mismatches between the textual semantics of $T$ and the visual context of $I$.

\subsection{Cross-modal Feature Alignment}
Given the $i$-th sample $(I_i, T_i)$, we utilize modality-specific encoders $E_v$ and $E_t$ to produce visual features $h_i^v \in \mathbb{R}^{d_v}$ and textual features $h_i^t \in \mathbb{R}^{d_t}$. 
To align features across modalities, we project both $h_i^v$ and $h_i^t$ into a shared latent space through learnable linear layers, obtaining $z_i^v$ and $z_i^t$ with the same dimensionality $d_z$.
We then adopt contrastive learning to align the projected features. The similarity score $s_{ij}$ between the $i$-th image and $j$-th text is defined by cosine similarity:
\begin{equation}
s_{ij} = \frac{(z_{i}^{v})^{\top} z_{j}^{t}}{\lVert z_{i}^{v} \rVert_{2}\,\lVert z_{j}^{t} \rVert_{2}}, \quad i,j \in \{1,\dots,B\},
\label{eq:cosine}
\end{equation}
where $B$ is the mini-batch size. The margin-based contrastive loss is calculated as:
\begin{equation}
\mathcal{L}_{\text{cont}} = \frac{1}{B} \sum_{i=1}^{B} \sum_{\substack{j=1\\ j\neq i}}^{B} \max\bigl(0,\; m + s_{ij} - s_{ii}\bigr),
\label{eq:contrastive_loss}
\end{equation}
where $m>0$ is a margin hyperparameter encouraging separation between positive and negative pairs. This objective maximizes the similarity of matched pairs while pushing apart mismatched ones, thereby enhancing cross-modal alignment.

\subsection{Generative Discrepancy Representation Module}
The Generative Discrepancy Representation Module (GDRM) captures implicit conflicts between the original text $T$ and the image $I$. The module first employs an MLLM, such as LLaVA-NEXT~\cite{liu2024llavanext}, to generate a textual description $\hat{T}$ from the image, ensuring that the description faithfully reflects visual semantics. To avoid sarcasm-related biases, the MLLM is restricted to image-only input, excluding any multimodal contextual cues. Consequently, $\hat{T}$ serves as an unbiased and context-independent representation of the image.

To quantify the inconsistency between the generated description $\hat{T}$ and the original text $T$, we compute three measures: semantic discrepancy, sentiment discrepancy, and visual-textual fidelity. 

The semantic discrepancy $d_{\text{sem}}$ measures the divergence in meaning between the original text and the generated description. We compute the cosine dissimilarity between the CLIP text embeddings of $T$ and $\hat{T}$.
The sentiment discrepancy $d_{\text{sen}}$ captures shifts in sentiment. We use a RoBERTa-based sentiment classifier~\cite{ott2019fairseq} to obtain sentiment probability distributions for both texts, calculating the discrepancy via $L_1$ distance. 
In addition, we measure visual-textual fidelity $d_{\text{fidelity}}$, defined as the alignment between the generated description and the image. This is quantified as the cosine similarity \(\cos(z_I, z_{\hat{T}})\) between the CLIP image embedding $z_I$ and the CLIP text embedding of $z_{\hat{T}}$. A lower $d_{\text{fidelity}}$ value signifies greater deviation between the generated text and the visual content.

The three measures are concatenated to form a discrepancy feature vector:
\begin{equation}
    D = d_{\text{sem}} \oplus d_{\text{sen}} \oplus d_{\text{fidelity}},
    \label{eq:discrepancy_vector}
\end{equation}
where $\oplus$ denotes concatenation. The vector $D$ is further processed by a Multilayer Perceptron (MLP) to obtain the final discrepancy representation $F_{D}$.

\subsection{Gated Multimodal Fusion \& Classification}
To effectively integrate textual, visual, and discrepancy-based features, we employ a gated fusion mechanism. This assigns learnable importance weights to each modality, enabling the model to adaptively focus on the most informative features. Given feature vectors from the text $F_T$, image $F_I$, and discrepancy $F_D$, we compute modality-specific weights using the following gating functions:
\begin{equation}
g_T = \sigma(W_T F_T), \quad g_I = \sigma(W_I F_I), \quad g_D = \sigma(W_D F_D),
\end{equation}
where $W_T$, $W_I$, $W_D$ are trainable parameters and $\sigma$ denotes the sigmoid activation function. The final fused representation is:
\begin{equation}
F_{\text{fused}} = g_T \odot F_T + g_I \odot F_I + g_D \odot F_D.
\end{equation}

For classification, we utilize four independent classifiers for each modality-specific feature vector, alongside the fused representation. These logits are concatenated to form a combined representation $\textit{logits}_{\text{all}}$. Subsequently, the concatenated logits are processed by an MLP to produce the final prediction $P_{\text{final}}$.

\subsection{Optimization Objective}
We train GDCNet to jointly optimize sarcasm classification and multimodal alignment. The objective consists of two components: a binary classification loss and a contrastive loss. 

Sarcasm detection is formulated as a binary classification problem, computed via the cross-entropy loss between the predicted $P_{\text{final}}$ and the ground truth label $y$. Given a batch of $N$ training samples, the classification loss is:
\begin{equation}
\mathcal{L}_{\text{BCE}} = 
-\frac{1}{N} \sum_{i=1}^{N} \left[
y_i \log P_{\text{final},i} + (1-y_i) \log (1 - P_{\text{final},i})
\right].
\label{eq:classification_loss}
\end{equation}

To enforce cross-modal consistency, we incorporate the contrastive loss $\mathcal{L}_{\text{cont}}$ (Eq.~\ref{eq:contrastive_loss}), which aligns paired image–text embeddings in the shared latent space. 
The final objective combines both terms, with a hyperparameter $\alpha$ controlling the trade-off between classification performance and multimodal alignment:
\begin{equation}
\mathcal{L} = \mathcal{L}_{\text{BCE}} + \alpha \mathcal{L}_{\text{cont}}.
\label{eq:final_loss}
\end{equation}

%% file: body/experiments.tex
\section{Experiments}
\subsection{Experiments Details}

\textit{\textbf{Dataset}}: We evaluate our approach on MMSD2.0 \cite{qin2023mmsd2}, a refined and reliable benchmark for multimodal sarcasm detection. Constructed upon the original MMSD dataset~\cite{cai2019multi}, it eliminates spurious cues and rectifies unreasonable annotations, thereby offering a more robust basis for evaluation.

\noindent\textit{\textbf{Baselines}}: To evaluate our approach, we compare it against competitive baselines in three categories. \textit{Text-only methods} include BiLSTM~\cite{zhou2016attention}, SMSD~\cite{xiong2019self}, and BERT~\cite{devlin2019bert}; \textit{image-only methods} consist of ResNet~\cite{he2016deep} and ViT~\cite{dosovitskiy2020image}; and \textit{multimodal methods} encompass InCrossMGs~\cite{liang2021cross}, HKE~\cite{liu2022sarcasm}, Multi-view CLIP~\cite{qin2023mmsd2}, DIP~\cite{wen2023dip}, TFCD~\cite{zhu2024tfcd}, MOBA~\cite{xie2024moba}, CofiPara~\cite{chen-etal-2024-cofipara}, and ADs~\cite{jana2025ads}.

\input{figure/results_comparison}

\noindent\textit{\textbf{Implementation Details}}:
We implement GDCNet using the CLIP backbone with feature dimensions of $512$ (text) and $768$ (image). For image caption generation, we employ LLaVA-Next-7B~\cite{liu2024llavanext}. Training is conducted on four NVIDIA RTX 4090 GPUs. The model is optimized via the Adam optimizer for 10 epochs with a batch size of 32. We employ a learning rate of $5\times10^{-4}$ for the specific task modules and $1\times10^{-6}$ for the CLIP backbone. To ensure training stability, we apply a weight decay of $0.05$ and gradient clipping with a maximum norm of $5.0$. For the final objective function, the contrastive loss margin is set to $m=0.2$, weighted by the hyperparameter $\alpha=0.1$.

\subsection{Main Results}
\label{sec:results}

Table~\ref{tab:performance} presents the comparative results, highlighting the efficacy of our proposed framework. 
GDCNet establishes a new state-of-the-art on the MMSD2.0 benchmark, consistently outperforming prior baselines in both accuracy and F1-score. 
This performance advantage stems from our novel modeling paradigm. In contrast to CofiPara, which relies on joint text-image sarcasm rationale generation, GDCNet decouples image description generation from textual input. This design effectively mitigates textual bias by producing neutral visual observations, thereby facilitating the precise quantification of sarcasm through semantic and sentiment divergence analysis.
Furthermore, the adaptive gated fusion mechanism explicitly incorporates cross-modal divergence features, preventing modality dominance and ensuring balanced multimodal integration.

\subsection{Ablation Study}
\label{sec:ablation}

To assess the contribution of GDRM and its specific components, we conduct ablations on three configurations: removing the entire GDRM (\textit{w/o GDRM}), and excluding semantic (\textit{w/o SemD}) or sentiment (\textit{w/o SenD}) discrepancies. Table~\ref{tab:ablation} details the results on the MMSD2.0 benchmark. 
Removing the entire GDRM yields the most significant performance drop, reducing Accuracy by 2.96\% and F1-score by 4.15\%. This underscores the critical role of explicit discrepancy modeling.
Specifically, the absence of SemD impairs the detection of literal contradictions in incongruent image-text pairs. Conversely, removing SenD limits the capacity to capture subtle sentiment polarity shifts in text-driven sarcasm. The complementarity of these components enables the full model to maintain balanced performance and superior robustness.

\begin{table}[b]
\centering
\footnotesize 
\renewcommand{\arraystretch}{1.1}
\caption{Ablation study on the MMSD2.0 dataset. The significant performance drops across all metrics upon removing GDRM or its sub-components validate their individual and collective importance.}
\setlength{\tabcolsep}{5pt} 
\begin{tabular}{lcccc}
\toprule
\textbf{Method} & \textbf{Acc. (\%)} & \textbf{P (\%)} & \textbf{R (\%)} & \textbf{F1 (\%)} \\
\midrule
\textbf{Full Model} & \textbf{87.38} & \textbf{83.39} & \textbf{89.51} & \textbf{86.34} \\
\midrule
\textit{- w/o GDRM} & 84.42  & 78.56  & 86.17  & 82.19  \\
\textit{- w/o SemD} & 86.23  & 80.27  & 87.09  & 83.54  \\
\textit{- w/o SenD} & 85.98  & 81.74  & 87.63  & 84.58  \\
\bottomrule
\end{tabular}
\label{tab:ablation}
\end{table}

\subsection{Comparison with LLM-based Methods}
To further evaluate the performance of GDCNet, we conduct a comparative analysis against various direct LLM-based methods for sarcasm detection on the MMSD2.0 dataset. As presented in Table \ref{tab:llm_comparison}, we evaluate prominent MLLMs such as LLaVA\cite{liu2023llava}, Qwen-VL\cite{bai2023qwenvlversatilevisionlanguagemodel}, and GPT-4o\cite{hurst2024gpt}, under both Zero-Shot and Chain-of-Thought (CoT) prompting strategies. 
GDCNet consistently outperforms all LLM-based baselines in all metrics. While CoT improves reasoning for some models, MLLMs still struggle with sarcasm detection, underscoring the advantage of GDCNet’s explicit discrepancy modeling for this challenging task.

\begin{table}[tb]
\centering
\footnotesize 
\renewcommand{\arraystretch}{1}
\caption{Comparison results for LLM-based methods and GDCNet on the MMSD2.0 dataset. GDCNet significantly outperforms general-purpose MLLMs, even when they utilize CoT prompting.}
\begin{tabular}{lcccc}
\toprule
\textbf{Method} & Acc. (\%) & P (\%) & R (\%) & F1 (\%) \\
\midrule
LLaVA (Zero-Shot) & 51.06 & 40.09 & 46.40 & 43.02 \\
LLaVA (CoT) & 48.69 & 40.93 & 65.17 & 50.28 \\
Qwen-VL (Zero-Shot) & 40.63 & 32.44 & 35.53 & 33.63 \\
Qwen-VL (CoT) & 58.86 & 56.82 & 58.67 & 57.26 \\
GPT-4o (Zero-Shot) & 71.07 & 79.52 & 71.07 & 70.24 \\
GPT-4o (CoT) & 74.26 & 65.81 & 72.68 & 68.92 \\
\midrule
\textbf{GDCNet(Ours)} & \textbf{87.38} & \textbf{83.39} & \textbf{89.51} & \textbf{86.34} \\
\bottomrule
\end{tabular}
\label{tab:llm_comparison}
\end{table}

\subsection{MLLM Contribution Analysis}
To investigate the impact of the MLLM on GDCNet's performance, we conducted an ablation study on the MMSD2.0 benchmark by substituting our image caption generator with BLIP-2\cite{li2023blip} and LLaVA-NEXT\cite{liu2024llavanext}. For each MLLM, we generate descriptions for all samples using a standardized prompt and train GDCNet with fixed hyperparameters for both configurations. As shown in Table~\ref{tab:mlm_comparison}, while BLIP-2 offers superior inference speed, LLaVA-NEXT generates significantly richer descriptions with higher semantic consistency. Crucially, richer semantic captions lead to consistently better downstream performance, highlighting the importance of caption quality in multimodal sarcasm detection.

\begin{table}[tb] 
\centering
\footnotesize 
\setlength{\tabcolsep}{4pt} 
\renewcommand{\arraystretch}{1} 
\caption{Performance and cost comparison of BLIP-2 and LLaVA-NEXT as caption generators in GDCNet. While BLIP-2 is faster, LLaVA-NEXT provides the detailed semantic grounding necessary for higher accuracy.}
\begin{tabular}{lccc|cc}
\toprule
\textbf{MLLM}  & Time (s) & Tokens & CLIP-S & Acc. (\%) & F1 (\%)\\
\midrule
BLIP-2        & \textbf{0.23} & 21.53 & 31.3 & 86.73  & 85.66  \\
LLaVA-NEXT  & 1.70 & \textbf{67.29}  & \textbf{49.2} & \textbf{87.38} & \textbf{86.34}\\
\bottomrule
\end{tabular}
\label{tab:mlm_comparison}
\end{table}

%% file: figure/results_comparison.tex
\begin{table}[t]
    \centering
    \footnotesize 
    \renewcommand{\arraystretch}{1}
    \caption{Performance comparison on MMSD2.0. Multimodal methods outperform unimodal ones, and GDCNet achieves the highest overall performance.}
    \begin{tabular}{l | c c c c}
        \toprule[1pt]
        \textbf{Method} & Acc.(\%) & P(\%) & R(\%) & F1(\%) \\
        \midrule
        \multicolumn{5}{c}{\textbf{Text-Only Methods}} \\
        \midrule
        BiLSTM \cite{zhou2016attention} & 72.48 & 68.02 & 68.08 & 68.05 \\
        SMSD \cite{xiong2019self} & 73.56 & 68.45 & 71.55 & 69.97 \\
        BERT \cite{devlin2019bert} & 76.52 & 74.48 & 73.09 & 73.78 \\
        \midrule
        \multicolumn{5}{c}{\textbf{Image-Only Methods}} \\
        \midrule
        ResNet \cite{he2016deep} & 65.50 & 61.17 & 54.39 & 57.58 \\
        ViT \cite{dosovitskiy2020image} & 72.02 & 65.26 & 74.83 & 69.72 \\
        \midrule
        \multicolumn{5}{c}{\textbf{Multi-Modal Methods}} \\
        \midrule
        InCrossMGs \cite{liang2021cross} & 79.83 & 75.82 & 78.01 & 76.90 \\
        HKE \cite{liu2022sarcasm} & 76.50 & 73.48 & 71.07 & 72.25 \\
        Multi-view CLIP \cite{qin2023mmsd2} & 85.14 & 80.33 & 88.24 & 84.09 \\
        DIP \cite{wen2023dip} & 84.63 & 84.17 & 85.20 & 84.68 \\
        TFCD \cite{zhu2024tfcd} & 86.54 & 82.46 & 87.95 & 84.31 \\
        MoBA \cite{xie2024moba} & 85.01 & 80.46 & 87.67 & 83.64 \\
        CofiPara \cite{chen-etal-2024-cofipara} & 85.66 & \textbf{85.79} & 85.43 & 85.61 \\
        ADs \cite{jana2025ads} & 85.60 & 85.28 & 85.60 & 85.41 \\
        \midrule
        \textbf{GDCNet} (\textbf{Ours}) & \textbf{87.38} & 83.39 & \textbf{89.51} & \textbf{86.34} \\
        \bottomrule[1pt]
    \end{tabular}
    \label{tab:performance}
\end{table}

%% file: body/conclusion.tex
\section{Conclusion}
In this paper, we propose GDCNet, a novel framework for multimodal sarcasm detection that effectively mitigates noise from LLM-generated sarcastic cues and explicitly models cross-modal incongruity. By leveraging LLM-generated image-grounded captions as cross-modal anchors, GDCNet captures cross-modal incongruity through fine-grained semantic and sentiment discrepancy modeling. Furthermore, an adaptive gated fusion module integrates visual, textual, and discrepancy signals, effectively alleviating modality dominance and reducing spurious correlations. Extensive experiments on the MMSD2.0 dataset demonstrate that GDCNet establishes a new state-of-the-art, consistently outperforming competitive baselines. These results underscore the potential of leveraging LLMs not merely as data generators but as structural guides for capturing subtle cross-modal incongruities, advancing the frontier of complex multimodal understanding tasks such as sarcasm detection.